# Granular Computing: An Augmented Scheme of Degranulation Through a Modified Partition Matrix

Kaijie Xu, Witold Pedrycz, *Fellow, IEEE*, Zhiwu Li, *Fellow, IEEE*, and Mengdao Xing, *Fellow, IEEE*

*Abstract*—As an important technology in artificial intelligence Granular Computing (GrC) has emerged as a new multi-disciplinary paradigm and received much attention in recent years. Information granules forming an abstract and efficient characterization of large volumes of numeric data have been considered as the fundamental constructs of GrC. By generating centroids (prototypes) and partition matrix, fuzzy clustering is a commonly encountered way of information granulation. As a reverse process of granulation, degranulation involves data reconstruction completed on a basis of the granular representatives (decoding information granules into numeric data). Previous studies have shown that there is a relationship between the reconstruction error and the performance of the granulation process. Typically, the lower the degranulation error is, the better performance of granulation process becomes. However, the existing methods of degranulation usually cannot restore the original numeric data, which is one of the important reasons behind the occurrence of the reconstruction error. To enhance the quality of reconstruction (degranulation), in this study, we develop an augmented scheme through modifying the partition matrix. By proposing the augmented scheme, we dwell on a novel collection of granulation-degranulation mechanisms. In the constructed approach, the prototypes can be expressed as the product of the dataset matrix and the partition matrix. Then, in the degranulation process, the reconstructed numeric data can be decomposed into the product of the partition matrix and the matrix of prototypes. Both the granulation and degranulation are regarded as generalized rotation between the data subspace and the prototype subspace with the partition matrix and the fuzzification factor. By modifying the partition matrix, the new partition matrix is constructed through a series of matrix operations. We offer a thorough analysis of the developed scheme. The experimental results are in agreement with the underlying conceptual framework. The results obtained on both synthetic and publicly available datasets are reported to show the enhancement of the data reconstruction performance thanks to the proposed method. It is pointed out that in some cases the reconstruction errors can be reduced to zero by using the proposed approach.

This work was supported in part by the National Natural Science Foundation of China under Grant Nos. 61672400 and 61971349. (*Corresponding author: Zhiwu Li.*)

K. Xu is with the School of Electro-Mechanical Engineering, Xidian University, Xi'an 710071, China and also with the Department of remote sensing science and technology, School of Electronic Engineering, Xidian University, Xi'an 710071, China. (e-mail: kjxu@stu.xidian.edu.cn).

W. Pedrycz is with the Department of Electrical and Computer Engineering, University of Alberta, Edmonton, AB T6R 2V4, Canada, the School of Electro-Mechanical Engineering, Xidian University, Xi'an 710071, China, Systems Research Institute, Polish Academy of Sciences, Warsaw, Poland and also with the Faculty of Engineering, King Abdulaziz University, Jeddah 21589, Saudi Arabia (e-mail: wpedrycz@ualberta.ca).

Z. Li is with the School of Electro-Mechanical Engineering, Xidian University, Xi'an 710071, China, and also with the Institute of Systems Engineering, Macau University of Science and Technology, Macau 999078, China (e-mail: zhwli@xidian.edu.cn).

M. Xing is with the National Laboratory of Radar Signal Processing, Xidian University, Xi'an 710071, China, Collaborative Innovation Center of Information Sensing and Understanding, Xidian University, Xi'an 710071, China, and also with the National Key Laboratory of Microwave Imaging Technology, Institute of Electronics, Chinese Academy of Sciences, Beijing 100190, China (e-mail: xmd@xidian.edu.cn).

## I. INTRODUCTION

Information granules [1] (fuzzy sets [2] and fuzzy relations, in particular) are the building blocks of fuzzy models, classifiers, and rule-based systems [3–7]. Information granules also have been considered to be the fundamental constructs of Granular Computing (GrC) [8–11]. Fuzzy clustering is one of the most commonly encountered approaches to construct information granules on the basis of experimental data. Fuzzy clustering methods basically focus on the abstraction of the original numeric data. Fuzzy C-Means (FCM), a fuzzy clustering method based on cost-function, has become a popular technique predominantly because of its simplicity and efficiency, and also due to the fact that it is based on a complete set of theoretical framework and mathematical theory [12–15]. Admittedly, information granules are formed based on the existing numeric evidence which gives rise to the ever-growing importance of various fuzzy clustering mechanisms and, to the relevance of the FCM [7], [16], in particular.

In the FCM algorithm, the structure in the dataset is expressed in terms of constructed prototypes (clusters) and partition matrix [17]. Subsequently, data are encoded to information granules with the aid of constructed prototypes and partition matrix. The reconstruction of information granules, usually referred to as a degranulation or decoding process, returns a numeric result [11]. The concept of granulation-degranulation plays an important role in GrC, just as fuzzification-defuzzification in fuzzy control systems, and analog-to-digital (A/D) as well as digital-to-analog (D/A) conversion systems in digital signal processing [7, 11, 18, 19].

So far, the topic of granulation-degranulation mechanism has not been intensively studied. The lack of a well-established body of knowledge opens up new opportunities but also calls for more investigations in this area. In [20–21], the degranulation (reconstruction) error is used as an evaluation index of the performance of the fuzzy clustering. Studies have shown that the reconstruction error depends on the granulation scheme and the selection of its parameters. In the FCM, these parameters refer to the fuzzification coefficient and the number of clusters (information granules), which determine the prototypes (clusters) and partition matrix. Typically, the lower the reconstruction error is, the better the performance of the clustering becomes. To reduce the reconstruction error, Hu *et al.* [11] make use of a linear transformation of partition matrix to obtain the high-level associations among the data being clustered, and then set up an adjustment mechanism modifying a localization of the prototypes; thus, the partition matrix and prototypes can be modified. Finally, the reconstruction error is reduced with the use of population-based optimization. In [24], a modified clonal mutation scheme is proposed to enhance the reconstruction performance. By adjusting the location of the prototypes, Galaviz *et al.* [21] also develop a cluster optimization algorithm. Izakian *et al.* [25] make use of the reconstruction error as a vehicle to select the partition matrix so as to determine the results of fuzzy clustering. In [26], the reconstruction error is also used as a criterion to assign an anomaly score to each subsequence in anomaly detection in time series data. Similarly, in [27], a DFuzzy method with deep learning based autoencoders for graph clustering is designed by minimizing the reconstruction error. By considering the reconstruction error as an indicator of the quality of constructed clusters, Casalino *et al.* [28] propose a dynamic incremental semi-supervised version of the standard FCM to adapt dynamically the number of clusters to capture adequately the evolving structure of streaming data. Recently, a new design method for a fuzzy radial basis function neural networks classifier is proposed in [29]. The algorithm is based on the conditional Fuzzy C-Means clustering algorithm and realized with the aid of auxiliary information, which extracted by the locally linear reconstruction algorithm, shown to be an effective classification approach. In summary, the reconstruction (degranulation) is an important evaluation index in fuzzy clustering. It exhibits a close relation with the partition and the prototypes, and in turn it can optimize the partition and the prototypes by minimizing the reconstruction error to enhance the performance of fuzzy clustering. Thus, optimizing the degranulation process is a very important and well-motivated issue.

All of the aforementioned studies expose some similarities: they mainly focus on using the optimization methods in the granulation process to enhance the granulation results. With the optimization methods, they can all achieve a reduction in the reconstruction error. However, they cannot usually make the information granules generate original data, which is one of the important reasons behind the occurrence of the reconstruction error. Thus, we will improve the performance of the degranulation (reduce the reconstruction error) from the perspective of restoring the original data.

In this paper, a novel reconstruction (degranulation) scheme is proposed. In the developed scheme, we build up a novel collection of models of granulation-degranulation mechanisms. We complete a thorough analysis of the relationship among the prototypes (clusters), partition matrix and the information granules. In the granulation process, the matrix of prototypes is expressed as the product of the dataset matrix and the partition matrix. While in the degranulation process, the reconstructed numeric data can also be decomposed into the product of the partition matrix and the matrix of prototypes. By building up a supervised learning mode of the granulation-degranulation based on the developed models, the partition matrix is optimized through a series of matrix operations. With the modified partition matrix, the original numeric data can be restored from the information granules which reduces the reconstruction error significantly. The experimental studies demonstrate that the proposed approach achieves better performance in comparison with the performance of the FCM-based degranulation. To the best of our knowledge, the idea of the proposed approach has not been considered in the previous studies.

This paper is organized as follows. The granulation-degranulation process is briefly reviewed in Section II. An augmented scheme of degranulation is dwelled upon in detail in Section III. Section IV includes an experimental setup and analysis of completed experiments. Section V covers some conclusions and identifies future research directions.

## II. GRANULATION-DEGRANULATION MECHANISMS

Let us consider that a numeric dataset $X \subset R^n$ comprising of $N$ objects with $n$ attributes (measurements) is clustered into $C$ groups. From a general point of view, fuzzy clustering aims to form information granules and to reveal a structure in the data. In the FCM clustering, we realize the minimization of the following objective function (performance index) [30], i.e.,

$$J_{FCM} = \sum_{i=1}^{N}\sum_{j=1}^{C}\mu_{ij}^m d_{ij}^2 = \sum_{i=1}^{N}\sum_{j=1}^{C}\mu_{ij}^m \|x_i - v_j\|^2$$
$$x_i = [x_1, x_2, \cdots, x_k, \cdots] \in R^{1 \times n}, \quad X = [x_1; x_2; \cdots; x_i; \cdots] \in R^{N \times n}$$
$$i = 1, 2, \cdots, N, \quad k = 1, 2, \cdots, n, \quad j = 1, 2, \cdots, C \quad (1)$$
s.t.
$$\sum_{j=1}^{C}\mu_{ij} = 1, \quad 0 < \sum_{i=1}^{N}\mu_{ij} < N$$

where $x_i$ is the $i$th datum, $v_j$ is the $j$th center (prototype) of the cluster, $\mu_{ij}$ is the degree of membership of individual $x_i$ to the cluster $j$, $m(m>1)$ is a fuzziness exponent (fuzziness coefficient), and $\|\cdot\|$ stands for some distance. While there is a substantial diversity as far as distance functions are concerned, here, we adhere to a weighted Euclidean distance taking on the following form [7]:

$$\|x_i - v_j\|^2 = \sum_{k=1}^{n}\frac{(x_{ik} - v_{jk})^2}{\sigma_k^2} \quad (2)$$

where $\sigma_k$ stands for a standard deviation of the $k$th variable. While not being computationally demanding, this type of distance is still quite flexible. The objective function shown above is minimized by iteratively updating the partition matrix $U$ and the prototypes $V$ [31]:

$$U = \left[\mu_1^T, \mu_2^T, \cdots, \mu_i^T, \cdots, \mu_N^T\right] \in R^{C \times N}$$
$$\mu_i = \left[\mu_{i1}, \mu_{i2}, \cdots, \mu_{ij}, \cdots, \mu_{iC}\right] \quad (3)$$
$$\mu_{ij} = \frac{1}{\sum_{t=1}^{C}(\frac{\|x_i - v_j\|}{\|x_i - v_t\|})^{\frac{2}{m-1}}}$$

$$V = \begin{bmatrix} v_1 \\ v_2 \\ \vdots \\ v_j \\ \vdots \\ v_C \end{bmatrix} \in R^{C \times n} \quad v_j = [v_{j1}, v_{j2}, \cdots, v_{jk}, \cdots, v_{jn}] \quad v_{jk} = \frac{\sum_{i=1}^{N}(\mu_{ij}^m x_{ik})}{\sum_{i=1}^{N}\mu_{ij}^m} \quad (4)$$

To facilitate the analysis, we build a diagonal matrix $\boldsymbol{\Phi}$ to decompose the matrix of prototypes:

$$\boldsymbol{\Phi} = \text{diag}\left\{\frac{1}{\sum_{j=1}^{N}\mu_{i1}^m}, \cdots, \frac{1}{\sum_{j=1}^{N}\mu_{ij}^m}, \cdots, \frac{1}{\sum_{j=1}^{N}\mu_{iC}^m}\right\} \in R^{C \times C} \quad (5)$$

then, we build a new model of the granulation, that is, the prototypes can be decomposed into:

$$V = \boldsymbol{\Gamma} X = \boldsymbol{\Phi} U^m X \quad (6)$$

where

$$U^m = \begin{bmatrix} \mu_{11}^m & \mu_{12}^m & \cdots & \mu_{1i}^m & \cdots & \mu_{1N}^m \\ \mu_{21}^m & \mu_{22}^m & \cdots & \mu_{2i}^m & \cdots & \mu_{2N}^m \\ \vdots & \vdots & \ddots & \vdots & \ddots & \vdots \\ \mu_{j1}^m & \mu_{j2}^m & \cdots & \mu_{ji}^m & \cdots & \mu_{jN}^m \\ \vdots & \vdots & \ddots & \vdots & \ddots & \vdots \\ \mu_{C1}^m & \mu_{C2}^m & \cdots & \mu_{Ci}^m & \cdots & \mu_{CN}^m \end{bmatrix} \quad (7)$$

It can be seen form (6) that the granulation is essentially a subspace rotation problem. With a generalized rotation matrix $\boldsymbol{\Gamma}$, the data subspace spanned by the column of $X$ can be rotated into the prototype subspace which is spanned by the column of $V$. A certain granulation-degranulation mechanism is inherently associated with fuzzy clustering [7]. The granulation-degranulation mechanisms can be organized in the two phases [11] as displayed in Fig.1:

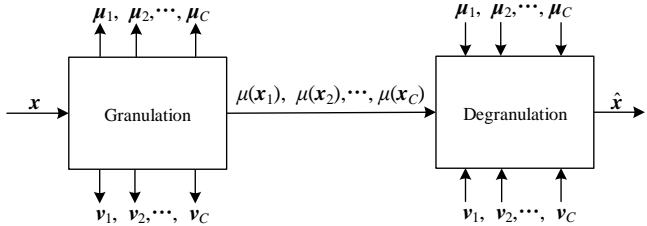

Fig. 1. Representation of the granulation-degranulation mechanisms.

Granulation of the data $x$ is made in terms of membership grades of the constructed information granules. In other words, the granulation mechanism returns a representation of any input data (pattern) $x$ expressed in terms of membership degrees, $\mu(x_1), \mu(x_2), \cdots \mu(x_C)$; the membership grades are computed by (3). Degranulation provides a reconstruction of the original data in terms of the prototypes and membership grades computed by (3) and (6), respectively. Formally, the reconstruction of data is determined by solving the following optimization problem where the minimized performance index is defined as follows:

$$\sum_{j=1}^{C} \mu_{ij}^m(x_i) \|x_i - v_j\|^2 \quad (8)$$

As a result, the minimized reconstruction error is obtained for the result of reconstruction expressed as follows:

$$\hat{x}_i = \frac{\sum_{j=1}^{C} \mu_{ij}^m v_j}{\sum_{j=1}^{C} \mu_{ij}^m} \quad (9)$$

The degranulation (reconstruction) error becomes a function that takes into account the prototypes of the clusters and the partition matrix; (9) is very similar to (4) used for prototype computing. Figs. 2 and 3 show an original data (glass identification dataset from the machine learning repository (http://archive.ics.uci.edu/ml) and the reconstructed data with the FCM-based degranulation. Obviously, there is a vast difference between the original data and the reconstructed data with the FCM algorithm, which is also an important reason leading to the reconstruction error. As we may observe from (9), the reconstructed dataset $\hat{X}$ is determined by $U^m$ and $V$. In order to reduce the reconstruction error, we consider optimizing $U^m$ and $V$.

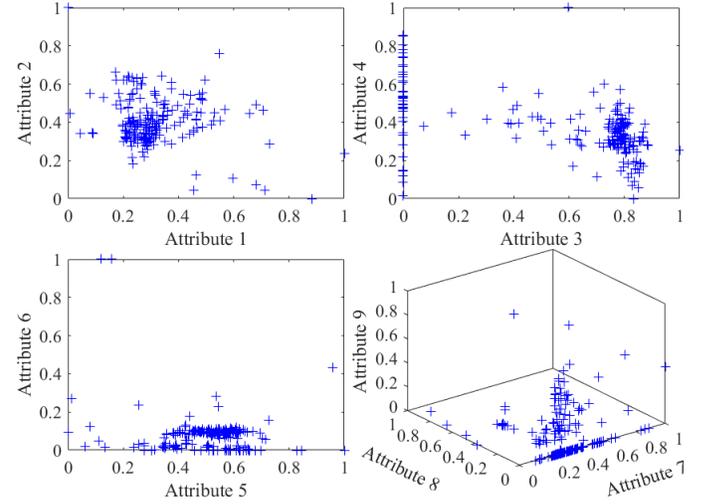

Fig. 2. Glass Identification dataset.

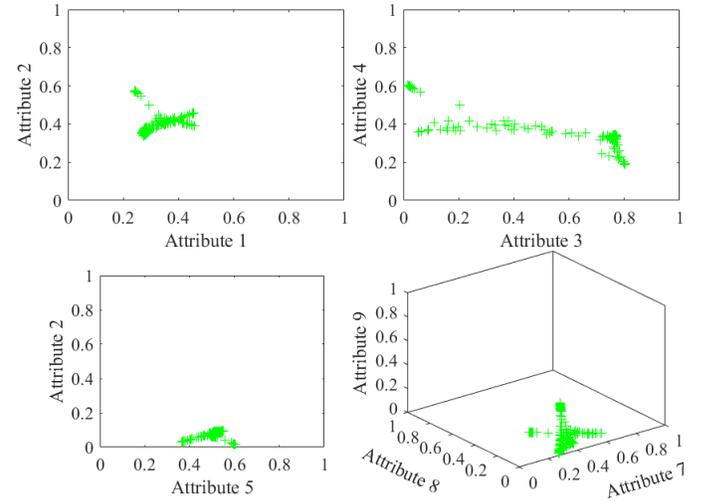

Fig. 3. Reconstructed data with the FCM-based degranulation.

III. AN AUGMENTED SCHEME OF DEGRANULATION

Although data can be reconstructed, there exists a difference between the reconstructed dataset and the original dataset, and the reconstruction error becomes nonzero. To enhance the performance of the degranulation process, in this paper we develop a novel approach to optimize the degranulation which can bring the reconstructed dataset closer to the original dataset (viz. restore the original dataset).

*A. Problem formulation*

To further explore the mathematical relationships between the partition matrix, the prototypes and the dataset, we build a novel model of the degranulation. That is, a matrix decomposition form of the reconstructed dataset $\hat{X}$, i.e.,

$$\hat{X} = \begin{bmatrix} \frac{\mu_{11}^m}{\sum_{j=1}^C \mu_{1j}^m} & \cdots & \frac{\mu_{1j}^m}{\sum_{j=1}^C \mu_{1j}^m} & \cdots \\ \vdots & \ddots & \vdots & \cdots \\ \frac{\mu_{i1}^m}{\sum_{j=1}^C \mu_{Nj}^m} & \cdots & \frac{\mu_{ij}^m}{\sum_{j=1}^C \mu_{ij}^m} & \cdots \\ \vdots & \ddots & \ddots & \ddots \end{bmatrix} V = \Omega V = \Theta \left[ U^m \right]^T V \quad (10)$$

$$i = 1, 2, \cdots, N; \quad j = 1, 2, \cdots, C$$

where

$$\Omega = \begin{bmatrix} \frac{\mu_{11}^m}{\sum_{j=1}^C \mu_{1j}^m} & \cdots & \frac{\mu_{1j}^m}{\sum_{j=1}^C \mu_{1j}^m} & \cdots \\ \vdots & \ddots & \vdots & \cdots \\ \frac{\mu_{i1}^m}{\sum_{j=1}^C \mu_{Nj}^m} & \cdots & \frac{\mu_{ij}^m}{\sum_{j=1}^C \mu_{ij}^m} & \cdots \\ \vdots & \ddots & \ddots & \ddots \end{bmatrix} = \Theta \left[ U^m \right]^T \quad (11)$$

$$i = 1, 2, \cdots, N; \quad j = 1, 2, \cdots, C$$

$$\Theta = \text{diag}\left\{ \frac{1}{\sum_{j=1}^C \mu_{1j}^m}, \cdots, \frac{1}{\sum_{j=1}^C \mu_{ij}^m}, \cdots, \frac{1}{\sum_{j=1}^C \mu_{Nj}^m} \right\} \in R^{N \times N} \quad (12)$$

It can be seen form (10) that the degranulation is also a subspace rotation problem. With a generalized rotation matrix $\Omega$, the prototype subspace spanned by the column of $V$ can be rotated into the data subspace which is spanned by the column of $X$.

Obviously, $U^m$ can be expressed as

$$\left[ U^m \right]^T = \begin{bmatrix} \left[ \sum_{t=1}^C (\frac{1}{\|x_1 - v_t\|})^{\frac{2}{m-1}} \right]^{-m} & 0 & 0 & 0 \\ 0 & \ddots & 0 & 0 \\ 0 & 0 & \left[ \sum_{t=1}^C (\frac{1}{\|x_i - v_t\|})^{\frac{2}{m-1}} \right]^{-m} & 0 \\ 0 & 0 & 0 & \ddots \end{bmatrix}$$

$$\begin{bmatrix} \|x_1 - v_1\|^{2\frac{-m}{m-1}} & \cdots & \|x_1 - v_j\|^{2\frac{-m}{m-1}} & \cdots \\ \vdots & \ddots & \vdots & \cdots \\ \|x_i - v_1\|^{2\frac{-m}{m-1}} & \cdots & \|x_i - v_j\|^{2\frac{-m}{m-1}} & \cdots \\ \vdots & \vdots & \vdots & \ddots \end{bmatrix} \quad (13)$$

$$i = 1, 2, \cdots, N, \; j = 1, 2, \cdots, C$$

As a result, $\hat{X}$ can be written in the following way

$$\hat{X} = \Lambda \Xi V \quad (14)$$

where

$$\Lambda = \text{diag}\left\{ \frac{\left[ \sum_{t=1}^C (\|x_1 - v_t\|^2)^{\frac{-1}{m-1}} \right]^{-m}}{\sum_{j=1}^C \mu_{1j}^m}, \cdots, \frac{\left[ \sum_{t=1}^C (\|x_i - v_t\|^2)^{\frac{-1}{m-1}} \right]^{-m}}{\sum_{j=1}^C \mu_{ij}^m}, \cdots \right\} = \Psi \Theta$$

$$i = 1, 2, \cdots, N; \quad j = 1, 2, \cdots, C \quad (15)$$

$$\Psi = \text{diag}\left\{ \left[ \sum_{t=1}^C (\|x_1 - v_t\|^2)^{\frac{-1}{m-1}} \right]^{-m}, \cdots, \left[ \sum_{t=1}^C (\|x_i - v_t\|^2)^{\frac{-1}{m-1}} \right]^{-m}, \cdots \right\} \quad (16)$$

$$i = 1, 2, \cdots, N; \quad j = 1, 2, \cdots, C$$

$$\Xi = \begin{bmatrix} \beta_1(v_1) & \beta_2(v_2) & \cdots & \beta_j(v_j) & \cdots & \beta_C(v_C) \end{bmatrix}$$

$$\beta_j(v_j) = \begin{bmatrix} \|x_1 - v_j\|^{2\frac{-m}{m-1}} & \cdots & \|x_i - v_j\|^{2\frac{-m}{m-1}} & \cdots & \|x_N - v_j\|^{2\frac{-m}{m-1}} \end{bmatrix}^T$$

$$i = 1, 2, \cdots, N; \quad j = 1, 2, \cdots, C \quad (17)$$

Furthermore, $\Omega = \Lambda \Xi$. We define $\Xi$ as a fuzzy curvature matrix of dataset $X$ while $\beta_j(v_j)$ is the curvature vector of the $j$-th prototype.

*B. Modification of the partition matrix*

Ideally, we wish to have $\hat{X} = X$. At this point, $X$ can be considered to be formed by $V$. Suppose that there exists a matrix $\hat{\Omega}$ between $V$ and the dataset $X$ that satisfies the following expression

$$\hat{\Omega} V = X \quad (18)$$

The modified $\hat{U}^m$ can be determined as follows

$$\hat{\Omega} V = X$$
$$\hat{\Omega} V V^T = X V^T \quad (19)$$
$$\hat{\Omega} = X V^T \left( V V^T \right)^{-1}$$

Since $\Omega = \Lambda \Xi$, $\Lambda$ being a diagonal matrix, in order to find a matrix $\hat{\Lambda}$ to satisfy $\hat{\Omega} = \hat{\Lambda} \Xi$, we construct a cost-function

$$F_\lambda = \sum_{i=1}^N \left\| \lambda_i \Xi_{(i)} - \hat{\Omega}_{(i)} \right\|^2 \quad (20)$$

where $\lambda_i$ is the $i$-th diagonal element, $\Xi_{(i)}$ and $\hat{\Omega}_{(i)}$ are the $i$-th row of the $\Xi$ and $\hat{\Omega}$, respectively. What we should do is to find a parameter $\hat{\lambda}_i$ to minimize the cost function $F_\lambda$. The expanded form of $F_\lambda$ reads as follows:

$$F_\lambda = \sum_{i=1}^N \left[ \lambda_i \Xi_{(i)} - \hat{\Omega}_{(i)} \right] \left[ \lambda_i^* \Xi_{(i)}^H - \hat{\Omega}_{(i)}^H \right]$$
$$= \sum_{i=1}^N \left[ \lambda_i \Xi_{(i)} \Xi_{(i)}^H \lambda_i^* - \lambda_i \Xi_{(i)} \hat{\Omega}_{(i)}^H - \lambda_i^* \hat{\Omega}_{(i)} \Xi_{(i)}^H + \hat{\Omega}_{(i)} \hat{\Omega}_{(i)}^H \right] \quad (21)$$

where $\lambda_i^*$ is conjugation of $\lambda_i$, and $H$ represents the conjugate transpose. Computing the partial derivative of $\lambda_i$ and make it equal to 0, we have

$$\frac{\partial F_\lambda}{\partial \lambda_i} = \Xi_{(i)} \Xi_{(i)}^H \lambda_i - \hat{\Omega}_{(i)} \Xi_{(i)}^H = 0 \quad (22)$$

As a result, $\hat{\lambda}_i$ can be expressed as

$$\hat{\lambda}_i = \left[ \Xi_{(i)} \Xi_{(i)}^H \right]^{-1} \hat{\Omega}_{(i)} \Xi_{(i)}^H \quad (23)$$

Thus, $\hat{\Lambda}$ can be written as

$$\hat{\Lambda} = \begin{bmatrix} \ddots & & \\ & \hat{\lambda}_{ii} & \\ & & \ddots \end{bmatrix}$$

$$\hat{\lambda}_{ii} = (\Xi_{ij}\Xi_{ij}^H)^{-1}(\hat{\Omega}_{ij}\Xi_{ij}^H) \quad (24)$$
$$i = 1,2,\cdots,N; \quad j = 1,2,\cdots,C$$

Another diagonal matrix $\hat{\Theta}$ is therefore obtained by

$$\hat{\Theta} = \Psi^{-1}\hat{\Lambda} \quad (25)$$

Finally, we use the idea of the total least squares (TLS) approximation [32]. To find $\hat{U}^m$, let

$$(\hat{\Theta} + \Delta\hat{\Theta})[U^m]^T = \hat{\Omega} + \Delta\hat{\Omega} \quad (26)$$

Specifically, we find a matrix $\hat{U}^m$ that minimizes error matrices $\Delta\hat{\Theta}$ and $\Delta\hat{\Omega}$ for $\hat{\Theta}$ and $\hat{\Omega}$, respectively. That is,

$$\begin{cases} (\hat{\Theta} + \Delta\hat{\Theta})[\hat{U}^m]^T = \hat{\Omega} + \Delta\hat{\Omega} \\ \min_{[U^m]^T,\Delta\hat{\Theta},\Delta\hat{\Omega}} \|[\Delta\hat{\Theta} \ \Delta\hat{\Omega}]\|_F^2 \end{cases} \quad (27)$$

where $[\Delta\hat{\Theta} \ \Delta\hat{\Omega}]$ is the augmented matrix with $\Delta\hat{\Theta}$ and $\Delta\hat{\Omega}$ side by side and $\|\bullet\|_F^2$ is the Frobenius norm. (27) can also be rewritten in the form

$$\begin{cases} [(\hat{\Theta} + \Delta\hat{\Theta}) \ (\hat{\Omega} + \Delta\hat{\Omega})]\begin{bmatrix}[\hat{U}^m]^T \\ -I_C\end{bmatrix} = 0 \\ \min_{[U^m]^T,\Delta\hat{\Theta},\Delta\hat{\Omega}} \|[\Delta\hat{\Theta} \ \Delta\hat{\Omega}]\|_F^2 \end{cases} \quad (28)$$

where $I_C$ is the $C \times C$ identity matrix, and $\begin{bmatrix}[\hat{U}^m]^T \\ -I_C\end{bmatrix}$ must be a matrix of rank $C$. Moreover, it is seen from (28) that $\begin{bmatrix}[\hat{U}^m]^T \\ -I_C\end{bmatrix}$ must lie in the right zero subspace of the augmented matrix $[(\hat{\Theta} + \Delta\hat{\Theta}) \ (\hat{\Omega} + \Delta\hat{\Omega})]$. Thus, $[(\hat{\Theta} + \Delta\hat{\Theta}) \ (\hat{\Omega} + \Delta\hat{\Omega})]$ has to be a rank-defect matrix that at least loses rank of $C$. If $\min_{[\hat{U}^m]^T,\Delta\hat{\Theta},\Delta\hat{\Omega}} \|[\Delta\hat{\Theta} \ \Delta\hat{\Omega}]\|_F^2$ is further considered, $[(\hat{\Theta} + \Delta\hat{\Theta}) \ (\hat{\Omega} + \Delta\hat{\Omega})]$ must lose rank of $C$, since if defect rank of $[(\hat{\Theta} + \Delta\hat{\Theta}) \ (\hat{\Omega} + \Delta\hat{\Omega})]$ is over $C$, then $\|[\Delta\hat{\Theta} \ \Delta\hat{\Omega}]\|_F^2$ necessarily increases. Our first goal is then to find $[\Delta\hat{\Theta} \ \Delta\hat{\Omega}]$ such that the rank of $[\hat{\Theta} \ \hat{\Omega}]$ decreases from $N+C$ to $N$. Define $[\Pi][\Sigma][G]^T$ as the singular value decomposition of the augmented matrix $[\hat{\Theta} \ \hat{\Omega}]$, where $\Pi$ and $G$ are two orthonormal matrices, and all the diagonal entries of diagonal matrix $\Sigma$ are constructed by all the singular values that are arranged in decreasing order. Then, we have

$$[\hat{\Theta} \ \hat{\Omega}] = [\Pi_{\hat{\Theta}} \ \Pi_{\hat{\Omega}}]\begin{bmatrix}\Sigma_{\hat{\Theta}} & 0 \\ 0 & \Sigma_{\hat{\Omega}}\end{bmatrix}[G_{\hat{\Theta}} \ G_{\hat{\Omega}}]^T \quad (29)$$

where $G$ is partitioned into two sub-parts corresponding to $\hat{\Theta}$ and $\hat{\Omega}$. The rank is reduced by setting some of the singular values to zero. Thus, if considering $\min_{[\hat{U}^m]^T,\Delta\hat{\Theta},\Delta\hat{\Omega}} \|[\Delta\hat{\Theta} \ \Delta\hat{\Omega}]\|_F^2$, we

should make

$$[\Delta\hat{\Theta} \ \Delta\hat{\Omega}] = \Pi_{\hat{\Omega}}\Sigma_{\hat{\Omega}}G_{\hat{\Omega}}^T \quad (30)$$

so that we have

$$[(\hat{\Theta} + \Delta\hat{\Theta}) \ (\hat{\Omega} + \Delta\hat{\Omega})] = \Pi_{\hat{\Theta}}\Sigma_{\hat{\Theta}}G_{\hat{\Theta}}^T \quad (31)$$

Hence, the right zero subspace of $[(\hat{\Theta} + \Delta\hat{\Theta}) \ (\hat{\Omega} + \Delta\hat{\Omega})]$ is spanned by $G_{\hat{\Omega}}$. The solution $\begin{bmatrix}[\hat{U}^m]^T \\ -I_C\end{bmatrix}$ must be spanned by $G_{\hat{\Omega}}$. This leads to

$$\begin{bmatrix}[\hat{U}^m]^T \\ -I_C\end{bmatrix} = G_{\hat{\Omega}}T \quad (32)$$

where $T \in R^{C \times C}$ is an appropriate matrix. Let $G_{\hat{\Omega}} = \begin{bmatrix}G_{\hat{\Omega}1} \\ G_{\hat{\Omega}2}\end{bmatrix}$, where $G_{\hat{\Omega}1} \in R^{N \times C}$, and $G_{\hat{\Omega}2} \in R^{C \times C}$, then (32) derives the following relations

$$[\hat{U}^m]^T = G_{\hat{\Omega}1}T \quad (33)$$
$$-I_C = G_{\hat{\Omega}2}T \quad (34)$$

The relations (33) directly follows

$$\hat{U}^m = [-G_{\hat{\Omega}1}G_{\hat{\Omega}21}^{-1}]^T \quad (35)$$

With the new $\hat{U}^m$, a new reconstructed dataset can be obtained. With the proposed approach (By modifying the partition matrix), the reconstructed dataset $\hat{X}$ is closer to the original dataset $X$. Fig. 4 shows the principle of the proposed scheme. Fig. 5 visualizes the reconstructed data of the glass identification dataset with the proposed scheme. It can be seen that the structure of the reconstructed data using the proposed scheme is much closer to the original dataset than the reconstructed data produced by the FCM-based degranulation.

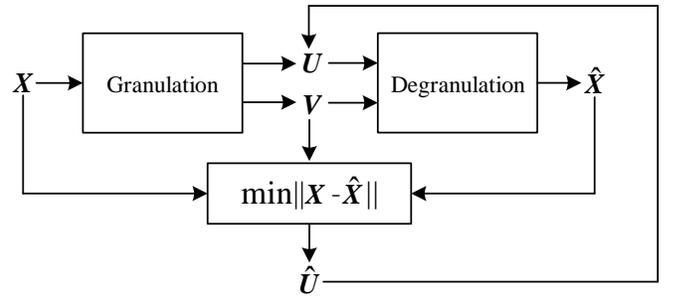

Fig. 4. An overall model: main functional processing phases.

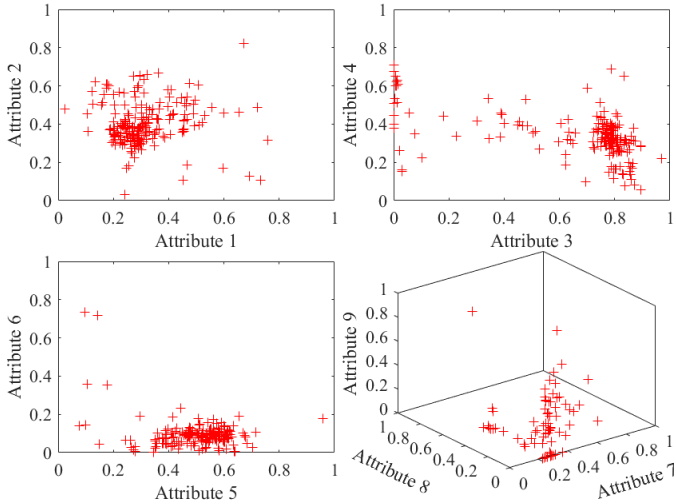

Fig. 5. Reconstructed data realized with the aid of the proposed method.

## IV. EXPERIMENTAL STUDIES

The following experiments are designed to evaluate the performance of the proposed approach and explore several different scenarios. The experiments are conducted for a variety of datasets using the FCM-based degranulation and the proposed method. A six-dimensional synthetic dataset is used, as seen in Fig. 6 in Section A. A number of publicly available datasets coming from the machine learning repository (http://archive.ics.uci.edu/ml) are also used. Data coming from the UCI machine learning repository [33] are commonly used as benchmarks [34]. All datasets are normalized to have zero mean and unit standard deviation. The reconstruction error is taken as the evaluation index

$$R_{error} = \frac{1}{N}\sum_{i=1}^{N}\|x_i - \hat{x}_i\|^2 \quad (36)$$

It should be noted that the normalized Euclidean distance is used in (36) to determine the discrepancy (distance) between the original $n$-dimensional entities (patterns), and the reconstructed ones.

We consider different values of the number of clusters $c$ and fuzzification coefficient $m$. We run the algorithms on each dataset with the number of clusters $c$ ranging from $C$ (here $C$ stands for the number of classes of each dataset) to $C+5$ [35]. The values of the fuzzification coefficient $m$ are taken from 1.2 to 3, with a step size of 0.2. The algorithms will be terminated once the following stopping condition is satisfied:

$$\max_{ij}\left|\mu_{ij}^{(k+1)} - \mu_{ij}^{(k)}\right| \leq 10^{-5} \quad (37)$$

To estimate the effectiveness of the proposed method, we used a 10-fold cross validation [22], [36] which is commonly used to estimate (and validate) the performance of granulation-degranulation models [21].

### A. Synthetic data

First, we report the results of reconstruction performance for an illustrative 6-D synthetic dataset, with a number of individuals 300 and three categories. To visualize and contrast the performance of the proposed and the FCM method, Figs. 6 to 8 show the original synthetic data and the reconstructed data with the FCM method and the proposed method, respectively.

One can notice that the shape of the reconstructed data with the proposed method is much closer to the original dataset than the FCM method. The best experimental results for this dataset are shown in Tables I. The results obtained for the 6-D synthetic dataset show that the proposed method is the clear winner in terms of the minimal reconstruction error.

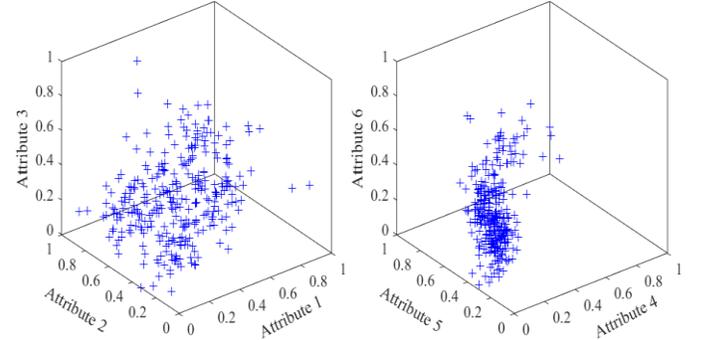

Fig. 6. Synthetic dataset.

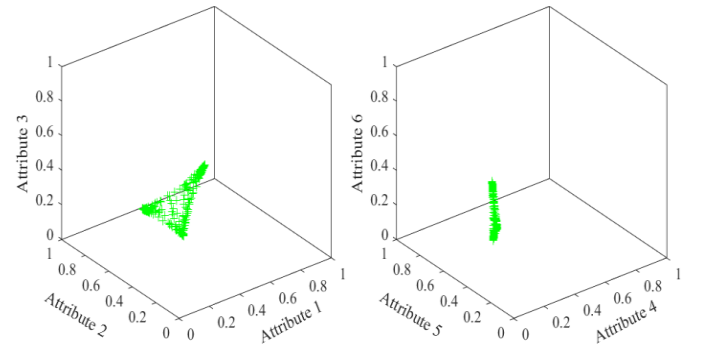

Fig. 7. Reconstructed data with the FCM-based degranulation.

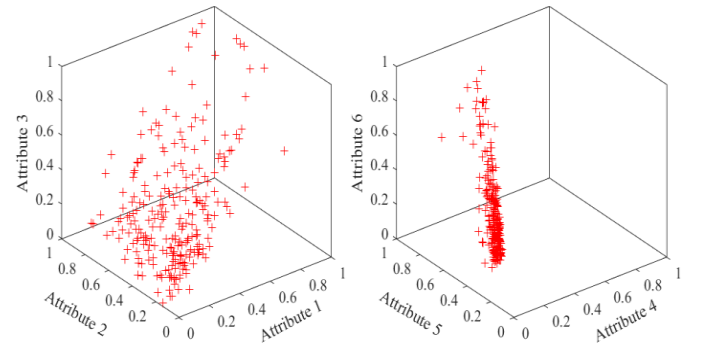

Fig. 8. Reconstructed data with the proposed method.

### B. Publicly available data

In this section, we use 11 publicly available datasets: vertebral column, glass identification, urban land cover, connectionist bench, climate model simulation crashes, breast cancer, qsar biodegradation, statlog (heart), adult, banknote authentication, and waveform database generator version2. The detailed introduction of these datasets can be found the UCI machine learning repository (http://archive.ics.uci.edu/ml). The reconstruction errors and standard derivations results of these indices obtained for each publicly available dataset are summarized in Tables II – VII and Fig. 9.

It is noticeable that the reconstruction error (both on the training and testing) of all the datasets are reduced by using the proposed approach. The improvement is about 25% on average and varies in-between a minimal improvement of 5% and 90% in the case of the most visible improvement. It is especially pointed out that in some cases the error can be reduced to zero.

The most notable result is the one obtained with the vertebral column dataset, where reconstruction errors of the training set and the test set reduce from 0.083 and 0.189 to 0.0002, respectively. The ones obtained with the glass identification and the urban land cover datasets also show good results, where in the former the reconstruction error is reduced by 90%, while in the latter the reconstruction error is reduced by about 30%. Although the reconstruction error of some dataset as waveform database generator version2 is slightly reduced, it is also improved by 5%.

The shortcoming of the proposed approach is an additional computational burden, due to the complex eigenvalue decomposition (EVD) and singular value decomposition (SVD) involved, especially for highly dimensional dataset.

Table I. Results of reconstruction error of 10-fold cross validation with all the protocols - synthetic dataset.

| Dataset | | synthetic dataset | | | | | | | | | |
|---|---|---|---|---|---|---|---|---|---|---|---|
| 10-Fold | | P1 | P2 | P3 | P4 | P5 | P6 | P7 | P8 | P9 | P10 |
| Parameters | $C$ | 8 | 8 | 8 | 3 | 8 | 8 | 8 | 8 | 8 | 8 |
| | $m$ | 1.4 | 1.2 | 1.4 | 1.4 | 1.2 | 1.4 | 1.2 | 1.2 | 1.4 | 1.2 |
| Train | FCM | 0.076 | 0.083 | 0.083 | 0.084 | 0.084 | 0.083 | 0.083 | 0.084 | 0.084 | 0.084 |
| | Proposed | 0.001 | 0.000 | 0.000 | 0.000 | 0.000 | 0.001 | 0.000 | 0.000 | 0.000 | 0.000 |
| Test | FCM | 0.173 | 0.186 | 0.173 | 0.193 | 0.201 | 0.205 | 0.200 | 0.173 | 0.207 | 0.180 |
| | Proposed | 0.001 | 0.000 | 0.000 | 0.000 | 0.000 | 0.000 | 0.000 | 0.000 | 0.001 | 0.000 |
| Total | FCM | 0.086 | 0.093 | 0.092 | 0.095 | 0.095 | 0.095 | 0.094 | 0.093 | 0.097 | 0.093 |
| | Proposed | 0.001 | 0.000 | 0.000 | 0.000 | 0.000 | 0.001 | 0.000 | 0.000 | 0.000 | 0.000 |
| Mean and standard deviation | | Train | | | | Test | | | | | |
| | | Mean | | Standard deviation | | | Mean | | Standard deviation | | |
| FCM | | 0.083 | | 0.002 | | | 0.189 | | 0.014 | | |
| Proposed | | 0.0002 | | 0.0002 | | | 0.0004 | | 0.0001 | | |

Table II. Results of reconstruction error of 10-fold cross validation with all the protocols - vertebral column dataset.

| Dataset | | Vertebral Column | | | | | | | | | |
|---|---|---|---|---|---|---|---|---|---|---|---|
| 10-Fold | | P1 | P2 | P3 | P4 | P5 | P6 | P7 | P8 | P9 | P10 |
| Parameters | $C$ | 8 | 8 | 8 | 8 | 8 | 8 | 8 | 8 | 8 | 8 |
| | $m$ | 1.2 | 1.2 | 1.2 | 1.2 | 1.2 | 1.2 | 1.4 | 1.2 | 1.2 | 1.4 |
| Train | FCM | 0.083 | 0.083 | 0.082 | 0.084 | 0.084 | 0.084 | 0.082 | 0.084 | 0.084 | 0.078 |
| | Proposed | 0.000 | 0.000 | 0.000 | 0.000 | 0.000 | 0.000 | 0.000 | 0.000 | 0.000 | 0.000 |
| Test | FCM | 0.197 | 0.201 | 0.210 | 0.203 | 0.189 | 0.176 | 0.195 | 0.176 | 0.154 | 0.186 |
| | Proposed | 0.000 | 0.000 | 0.000 | 0.000 | 0.000 | 0.000 | 0.000 | 0.000 | 0.000 | 0.001 |
| Total | FCM | 0.095 | 0.095 | 0.095 | 0.096 | 0.094 | 0.093 | 0.094 | 0.093 | 0.091 | 0.089 |
| | Proposed | 0.000 | 0.000 | 0.000 | 0.000 | 0.000 | 0.000 | 0.000 | 0.000 | 0.000 | 0.000 |
| Mean and standard deviation | | Train | | | | Test | | | | | |
| | | Mean | | Standard deviation | | | Mean | | Standard deviation | | |
| FCM | | 0.083 | | 0.002 | | | 0.189 | | 0.016 | | |
| Proposed | | 0.0002 | | 0.0001 | | | 0.0003 | | 0.0001 | | |

Table III. Results of reconstruction error of 10-fold cross validation with all the protocols - glass identification dataset.

| Dataset | | Glass Identification | | | | | | | | | |
|---|---|---|---|---|---|---|---|---|---|---|---|
| 10-Fold | | P1 | P2 | P3 | P4 | P5 | P6 | P7 | P8 | P9 | P10 |
| Parameters | $C$ | 12 | 12 | 12 | 12 | 12 | 12 | 12 | 12 | 12 | 12 |
| | $m$ | 1.2 | 1.2 | 1.2 | 1.2 | 1.2 | 1.2 | 1.2 | 1.2 | 1.2 | 1.2 |
| Train | FCM | 0.115 | 0.114 | 0.110 | 0.108 | 0.111 | 0.113 | 0.112 | 0.107 | 0.108 | 0.111 |
| | Proposed | 0.006 | 0.006 | 0.006 | 0.005 | 0.006 | 0.006 | 0.006 | 0.005 | 0.005 | 0.005 |
| Test | FCM | 0.154 | 0.082 | 0.155 | 0.146 | 0.136 | 0.123 | 0.145 | 0.151 | 0.163 | 0.129 |
| | Proposed | 0.011 | 0.010 | 0.010 | 0.014 | 0.011 | 0.011 | 0.007 | 0.016 | 0.029 | 0.018 |
| Total | FCM | 0.119 | 0.111 | 0.114 | 0.112 | 0.114 | 0.114 | 0.115 | 0.111 | 0.114 | 0.113 |
| | Proposed | 0.006 | 0.006 | 0.006 | 0.006 | 0.006 | 0.006 | 0.006 | 0.007 | 0.008 | 0.006 |
| Mean and standard deviation | | Train | | | | | Test | | | | |
| | | Mean | | Standard deviation | | | Mean | | Standard deviation | | |
| FCM | | 0.111 | | 0.003 | | | 0.138 | | 0.023 | | |
| Proposed | | 0.006 | | 0.000 | | | 0.014 | | 0.006 | | |

Table IV. Results of reconstruction error of 10-fold cross validation with all the protocols - urban land cover dataset.

| Dataset | | Urban Land Cover | | | | | | | | | |
|---|---|---|---|---|---|---|---|---|---|---|---|
| 10-Fold | | P1 | P2 | P3 | P4 | P5 | P6 | P7 | P8 | P9 | P10 |
| Parameters | $C$ | 7 | 7 | 7 | 7 | 7 | 7 | 7 | 7 | 7 | 7 |
| | $m$ | 1.2 | 1.2 | 1.2 | 1.2 | 1.2 | 1.2 | 1.2 | 1.2 | 1.2 | 1.2 |
| Train | FCM | 0.329 | 0.329 | 0.328 | 0.329 | 0.329 | 0.328 | 0.328 | 0.329 | 0.330 | 0.331 |
| | Proposed | 0.232 | 0.232 | 0.231 | 0.233 | 0.233 | 0.231 | 0.232 | 0.228 | 0.233 | 0.236 |
| Test | FCM | 0.882 | 0.841 | 0.876 | 0.859 | 0.869 | 0.857 | 0.872 | 0.875 | 0.877 | 0.837 |
| | Proposed | 0.603 | 0.630 | 0.630 | 0.622 | 0.624 | 0.615 | 0.618 | 0.624 | 0.618 | 0.604 |
| Total | FCM | 0.384 | 0.380 | 0.383 | 0.382 | 0.382 | 0.380 | 0.382 | 0.383 | 0.384 | 0.382 |
| | Proposed | 0.269 | 0.271 | 0.271 | 0.271 | 0.271 | 0.269 | 0.270 | 0.268 | 0.271 | 0.272 |
| Mean and standard deviation | | Train | | | | | Test | | | | |
| | | Mean | | Standard deviation | | | Mean | | Standard deviation | | |
| FCM | | 0.329 | | 0.001 | | | 0.864 | | 0.015 | | |
| Proposed | | 0.232 | | 0.002 | | | 0.619 | | 0.009 | | |

Table V. Results of reconstruction error of 10-fold cross validation with all the protocols - connectionist bench dataset.

| Dataset | | Connectionist Bench | | | | | | | | | |
|---|---|---|---|---|---|---|---|---|---|---|---|
| 10-Fold | | P1 | P2 | P3 | P4 | P5 | P6 | P7 | P8 | P9 | P10 |
| Parameters | $C$ | 7 | 7 | 7 | 7 | 7 | 7 | 7 | 7 | 7 | 7 |
| | $m$ | 1.2 | 1.2 | 1.2 | 1.2 | 1.2 | 1.2 | 1.2 | 1.2 | 1.2 | 1.2 |
| Train | FCM | 0.434 | 0.432 | 0.433 | 0.433 | 0.430 | 0.437 | 0.435 | 0.427 | 0.437 | 0.429 |
| | Proposed | 0.394 | 0.386 | 0.398 | 0.412 | 0.391 | 0.407 | 0.398 | 0.390 | 0.399 | 0.393 |
| Test | FCM | 0.989 | 1.105 | 1.014 | 1.085 | 1.084 | 0.968 | 1.002 | 1.094 | 1.019 | 1.109 |
| | Proposed | 0.969 | 1.063 | 1.015 | 0.983 | 1.005 | 0.830 | 0.889 | 1.029 | 0.941 | 1.023 |
| Total | FCM | 0.490 | 0.500 | 0.492 | 0.499 | 0.496 | 0.490 | 0.493 | 0.494 | 0.496 | 0.498 |
| | Proposed | 0.452 | 0.454 | 0.460 | 0.469 | 0.453 | 0.450 | 0.447 | 0.455 | 0.454 | 0.456 |
| Mean and standard deviation | | Train | | | | | Test | | | | |
| | | Mean | | Standard deviation | | | Mean | | Standard deviation | | |
| FCM | | 0.433 | | 0.003 | | | 1.047 | | 0.054 | | |
| Proposed | | 0.397 | | 0.008 | | | 0.975 | | 0.071 | | |

Table VI. Results of reconstruction error of 10-fold cross validation with all the protocols - climate model simulation crashes dataset.

| Dataset | | Climate Model Simulation Crashes | | | | | | | | | |
|---|---|---|---|---|---|---|---|---|---|---|---|
| 10-Fold | | P1 | P2 | P3 | P4 | P5 | P6 | P7 | P8 | P9 | P10 |
| Parameters | C | 7 | 7 | 7 | 7 | 6 | 7 | 7 | 7 | 7 | 4 |
| | m | 1.2 | 1.2 | 1.2 | 1.2 | 1.2 | 1.2 | 1.2 | 1.2 | 1.2 | 1.2 |
| Train | FCM | 0.192 | 0.192 | 0.192 | 0.192 | 0.192 | 0.192 | 0.192 | 0.192 | 0.192 | 0.192 |
| | Proposed | 0.158 | 0.174 | 0.160 | 0.158 | 0.165 | 0.160 | 0.157 | 0.157 | 0.161 | 0.176 |
| Test | FCM | 0.465 | 0.469 | 0.474 | 0.468 | 0.477 | 0.471 | 0.467 | 0.463 | 0.467 | 0.511 |
| | Proposed | 0.433 | 0.408 | 0.433 | 0.435 | 0.432 | 0.421 | 0.430 | 0.412 | 0.415 | 0.451 |
| Total | FCM | 0.220 | 0.220 | 0.220 | 0.220 | 0.221 | 0.220 | 0.220 | 0.219 | 0.220 | 0.224 |
| | Proposed | 0.185 | 0.197 | 0.187 | 0.186 | 0.192 | 0.186 | 0.185 | 0.183 | 0.186 | 0.204 |
| Mean and standard deviation | | Train | | | | | Test | | | | |
| | | Mean | | Standard deviation | | | Mean | | Standard deviation | | |
| FCM | | 0.192 | | 0.000 | | | 0.473 | | 0.014 | | |
| Proposed | | 0.163 | | 0.007 | | | 0.427 | | 0.013 | | |

Table VII. Results of reconstruction error of 10-fold cross validation with all the protocols - breast cancer dataset.

| Dataset | | Breast Cancer | | | | | | | | | |
|---|---|---|---|---|---|---|---|---|---|---|---|
| 10-Fold | | P1 | P2 | P3 | P4 | P5 | P6 | P7 | P8 | P9 | P10 |
| Parameters | C | 7 | 7 | 7 | 7 | 7 | 7 | 7 | 7 | 7 | 7 |
| | m | 1.2 | 1.4 | 1.4 | 1.4 | 1.4 | 1.4 | 1.4 | 1.4 | 1.4 | 1.4 |
| Train | FCM | 0.140 | 0.139 | 0.139 | 0.140 | 0.138 | 0.139 | 0.140 | 0.140 | 0.140 | 0.139 |
| | Proposed | 0.109 | 0.101 | 0.151 | 0.156 | 0.102 | 0.149 | 0.104 | 0.102 | 0.099 | 0.108 |
| Test | FCM | 0.353 | 0.346 | 0.346 | 0.344 | 0.351 | 0.369 | 0.352 | 0.343 | 0.341 | 0.340 |
| | Proposed | 0.229 | 0.300 | 0.235 | 0.286 | 0.338 | 0.245 | 0.319 | 0.292 | 0.272 | 0.248 |
| Total | FCM | 0.161 | 0.160 | 0.160 | 0.161 | 0.160 | 0.163 | 0.161 | 0.160 | 0.160 | 0.159 |
| | Proposed | 0.140 | 0.139 | 0.139 | 0.140 | 0.138 | 0.139 | 0.140 | 0.140 | 0.140 | 0.139 |
| Mean and standard deviation | | Train | | | | | Test | | | | |
| | | Mean | | Standard deviation | | | Mean | | Standard deviation | | |
| FCM | | 0.139 | | 0.001 | | | 0.349 | | 0.009 | | |
| Proposed | | 0.118 | | 0.024 | | | 0.276 | | 0.037 | | |

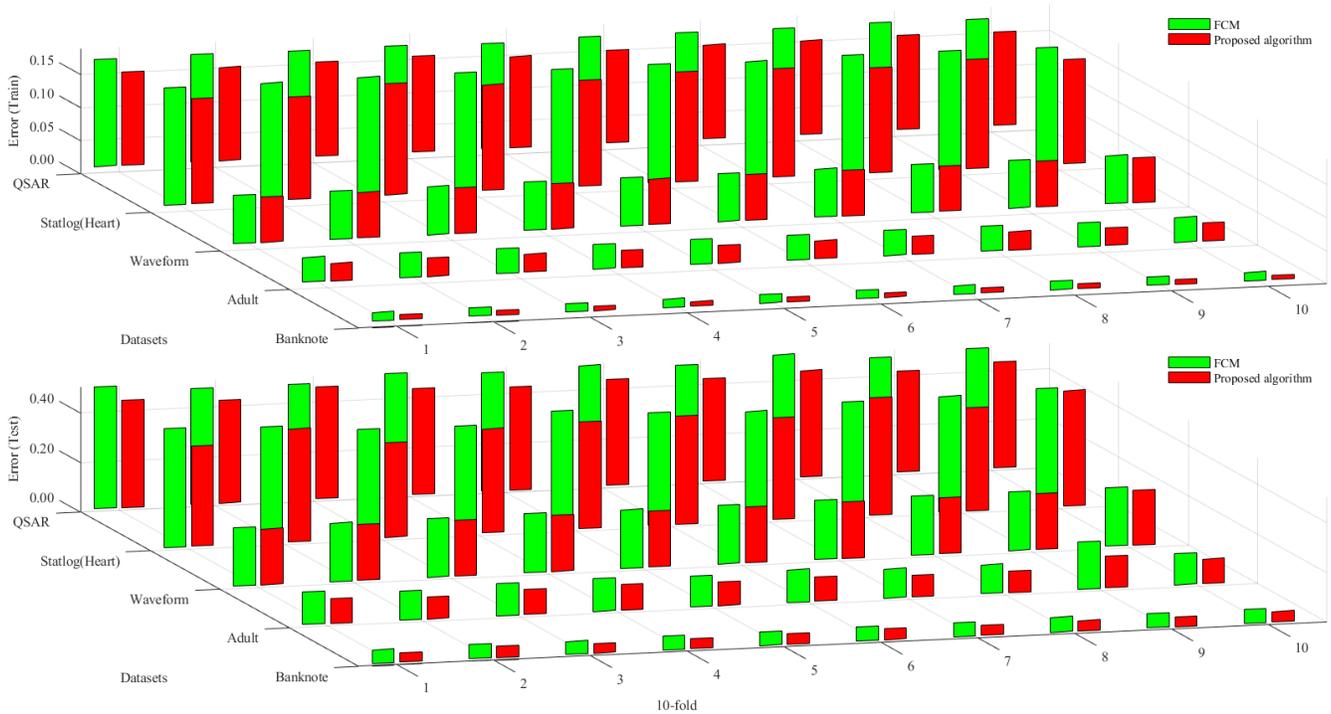

Fig. 9. Results of reconstruction error of 10-fold cross validation with all the protocols

## V. Conclusions

In this research, we propose a novel method to enhance the performance of the degranulation. During the design process, we build up a novel set of mathematical models of granulation-degranulation mechanisms so as to clearly present the relationship between the reconstructed dataset and the prototypes matrix. The partition matrix is modified by building up a supervised learning mode of the granulation-degranulation based on the developed models. After modifying the partition matrix, the reconstructed dataset is much closer to the original dataset, which significantly reduces the reconstruction error. We conduct a theoretical analysis and offer a comprehensive suite of experiments. Both the theoretical and experimental results are presented to validate the proposed method. To the best of our knowledge, this research scheme is first proposed that steadily improves the performance of degranulation. Unfortunately, the proposed method involves eigenvalue decomposition (EVD) and singular value decomposition (SVD), which involves some extra computing overhead.

The paper presents a way to modify the partition matrix to reduce the degranulation (reconstruction) error based on the proposed models of granulation-degranulation mechanisms, however, it is not the only way. Thus, the proposed models open a specific way for restoring the original dataset (reducing the reconstruction error) research and pose a much general problem as to the reduction of computational complexity. In addition, future work also includes the study of the model constructed in this paper combined with Petri nets [37] to reduce the degranulation error.

## References

[1] W. Pedrycz, "Granular computing for data analytics: a manifesto of human-centric computing," *IEEE/CAA Journal of Automatica Sinica,* vol. 5, no. 6, pp. 1025–1034, Nov. 2018.

[2] L. A. Zadeh, "Fuzzy sets," *Information & Control,* vol. 8, no. 3, pp. 338–353, Jan. 1965.

[3] F. Rogai, C. Manfredi, and L. Bocchi, "Metaheuristics for specialization of a segmentation algorithm for ultrasound images," *IEEE Transactions on Evolutionary Computation,* vol. 20, no. 5, pp. 730–741, Jan. 2016.

[4] J. Kerr-Wilson and W. Pedrycz, "Design of rule-based models through information granulation," *Expert Systems with Applications,* vol. 46, no. 2, pp. 274–285, Mar. 2016.

[5] X. C. Hu, W. Pedrycz, and O. Castillo, "Fuzzy rule-based models with interactive rules and their granular generalization," *Fuzzy Sets and Systems,* vol. 307, pp. 1–28, Jan. 2017.

[6] K. Neshatian, M. J. Zhang, and P. Andreae, "A filter approach to multiple feature construction for symbolic learning classifiers using genetic programming," *IEEE Transactions on Evolutionary Computation,* vol. 16, no. 5, pp. 645–661. Oct. 2012.

[7] W. Pedrycz and J. V. D. Oliveira, "A development of fuzzy encoding and decoding through fuzzy clustering," *IEEE Transactions on Instrumentation and Measurement,* vol. 57, no. 4, pp. 829–837, Mar. 2008.

[8] W. Pedrycz, "From fuzzy models to granular fuzzy models," *International Journal of Computational Intelligence Systems,* vol. 9, no. 1, pp. 35–42, Apr. 2016.

[9] G. Wilke and E. Portmann, "Granular computing as a basis of human-data interaction: a cognitive city use case," *Granular Computing,* vol. 1, no. 3, pp. 181–197, Sep. 2016.

[10] M. A. Sanchez, O. Castillo, and J. R. Castro, "Information granule formation via the concept of uncertainty-based information with interval type-2 fuzzy sets representation and Takagi–Sugeno–Kang consequents optimized with Cuckoo search," *Applied Soft Computing,* vol. 27, pp. 602–609, Feb. 2015.

[11] X. C. Hu, W. Pedrycz, G. H. Wu, and X. M. Wang, "Data reconstruction with information granules: an augmented method of fuzzy clustering," *Applied Soft Computing,* vol. 55, pp. 523–532, Jun. 2017.

[12] J. C. Bezdek, "Pattern recognition with fuzzy objective function algorithms," *Kluwer Academic Publishers,* pp. 203–239, Jan. 1981.

[13] C. Lian, R. Su, and T. Denœux, "Dissimilarity metric learning in the belief function framework," *IEEE Transactions on Fuzzy Systems,* vol. 24, no. 6, pp. 1555–1564, Mar. 2016.

[14] N. Ghadiri, M. Ghaffari, and M. A. Nikbakht, "Big FCM: fast, precise and scalable FCM on Hadoop," *Future Generation Computer Systems,* vol. 77, pp. 29–39, Dec. 2017.

[15] O. Castillo, L. Cervantes, and J. Soria, "A generalized type-2 fuzzy granular approach with applications to aerospace," *Information Sciences,* vol. 354: pp. 165–177, Aug. 2016.

[16] E. Rubio, O. Castillo, F. Valdez, P. Melin, C. I. Gonzalez, and G. Martinez, "An extension of the fuzzy possibilistic clustering algorithm using type-2 fuzzy logic techniques," *Advances in Fuzzy Systems,* doi: 10.1155/2017/7094046, 2017.

[17] D. Kumar, J. C. Bezdek, and M. Palaniswami, "A hybrid approach to clustering in big data," *IEEE Transactions on Cybernetics,* vol. 46, no. 10, pp. 2372–2385, Sep. 2015.

[18] K. J. Xu, W. K. Nie, D. Z. Feng, X. J. Chen, and D. Y. Fang, "A multi-direction virtual array transformation algorithm for 2D DOA estimation," *Signal Processing,* vol. 125, no. C, pp. 122–133, Aug. 2016.

[19] K. J. Xu, W. Pedrycz, Z. W. Li, and W. K. Nie, "High-accuracy signal subspace separation algorithm based on gaussian kernel," *IEEE Transactions on Industrial Electronics,* vol. 66, no. 1, pp. 491–499, Jan. 2019.

[20] X. B. Zhu, W. Pedrycz, and Z. W. Li, "Fuzzy clustering with nonlinearly transformed data," *Applied Soft Computing,* vol. 61, pp. 364–376, Jan. 2017.

[21] O. F. R. Galaviz and W. Pedrycz, "Enhancement of the classification and reconstruction performance of fuzzy c-means with refinements of prototypes," *Fuzzy Sets and Systems,* vol. 318, pp. 80–99, Jul. 2017.

[22] D. Graves and W. Pedrycz, "Kernel-based fuzzy clustering and fuzzy clustering: a comparative experimental study," *Fuzzy Sets and Systems,* vol. 161, no. 4, pp. 522–543, Feb. 2010.

[23] H. Izakian and W. Pedrycz, "Anomaly detection and characterization in spatial time series data: a cluster-centric approach," *IEEE Transactions on Fuzzy Systems,* vol. 22, no. 6, pp. 1612–1624, Dec. 2014.

[24] R. Liu, B. Zhu, R. Bian, Y. Ma, and L. Jiao. "Dynamic local search based immune automatic clustering algorithm and its applications," *Applied Soft Computing,* vol. 27, no. C, pp. 250–268, Feb. 2015.

[25] H. Izakian, W. Pedrycz, and I. Jamal, "Clustering spatiotemporal data: an augmented Fuzzy C-Means," *IEEE Transactions on Fuzzy Systems,* vol. 21, no. 5, pp. 855–868, Oct. 2013.

[26] I. Hesam and W. Pedrycz. "Anomaly detection in time series data using a fuzzy c-means clustering," *2013 Joint IFSA World Congress and NAFIPS Annual Meeting (IFSA/NAFIPS),* Edmonton, AB, Canada, Jun. 2013.

[27] V. Bhatia and R. Rinkle, "Dfuzzy: a deep learning-based fuzzy clustering model for large graphs," *Knowledge and Information Systems,* vol. 57, no. 1, pp. 159–181, Oct. 2018.

[28] G. Casalino, C. Giovanna, and M. Corrado, "Incremental adaptive semi-supervised fuzzy clustering for data stream classification," *2018 IEEE Conference on Evolving and Adaptive Intelligent Systems (EAIS),* Rhodes, Greece, Jun. 2018.

[29] S. B. Roh, S. K. Oh, W. Pedrycz, "Design methodology for radial basis function neural networks classifier based on locally linear reconstruction and conditional Fuzzy C-Means clustering," *International Journal of Approximate Reasoning,* vol. 106, pp. 228–243, Mar. 2019.

[30] Q. Wen, L. Yu, Y. Wang, and W. Wang, "Improved FCM algorithm based on the initial clustering center selection," in *Proc. International Conference on Consumer Electronics, Communications and Networks IEEE,* pp. 351–354, Xianning, China, Jan. 2014.

[31] J. M. L and H. W. Lewis, "Fuzzy clustering algorithms-review of the applications," *IEEE International Conference on Smart Cloud IEEE,* pp. 282-288, New York, USA, Dec. 2016.

[32] G. H. Gene and C. F. Van Loan, "Matrix computations (3rd ed.)," *Johns Hopkins University Press Baltimore,* MD. USA, 1996.

[33] J. S. Aguilar-Ruiz, R. Giraldez, J. C. Riquelme, Natural encoding for evolutionary supervised learning," *IEEE Transactions on Evolutionary Computation,* vol. 11, no. 4, pp. 466–479, Jul. 2007.

[34] E. Hullermeier, M. Rifqi, S. Henzgen, and R. Senge, "Comparing fuzzy partitions: a generalization of the rand index and related measures," *IEEE Transactions on Fuzzy Systems,* vol. 20, no. 3, pp. 546–556, Jun. 2012.

[35] K. J. Xu, W. Pedrycz, Z. W. Li, and W. K. Nie, "Constructing a virtual space for enhancing the classification performance of fuzzy clustering," *IEEE Transactions on Fuzzy Systems,* vol. 27, no. 9, pp.1779–1792, Sep. 2019.

[36] E. Adeli, K. H. Thung, and L. An, "Semi-supervised discriminative classification robust to sample-outliers and feature-noises," *IEEE transactions on pattern analysis and machine intelligence,* vol. 41, no. 2, p. 515–522, Feb. 2019.

[37] Z. Y. Ma, Z. W. Li, and A. Giua, "Marking estimation in a class of time labelled Petri nets," *IEEE Transactions on Automatic Control*, doi: 10.1109/tac.2019.2907413, 2019.